\documentclass[10pt,twocolumn,letterpaper]{article}

\usepackage{iccv}
\usepackage{times}
\usepackage{epsfig}
\usepackage{graphicx}
\usepackage{amsmath}
\usepackage{amssymb}
\usepackage{multirow,placeins,float}
 
\usepackage[pagebackref=true,breaklinks=true,letterpaper=true,colorlinks,bookmarks=false]{hyperref}

\iccvfinalcopy 

\def\iccvPaperID{1077} 

\ificcvfinal\fi

\author{Oran Gafni,
\quad Lior Wolf\\
Facebook AI Research and Tel-Aviv University\\
{\tt\small \{oran,wolf\}@fb.com}
\and Yaniv Taigman\\
Facebook AI Research\\
{\tt\small yaniv@fb.com}}

\usepackage{array}
\newcolumntype{x}[1]{>{\centering\arraybackslash\hspace{0pt}}p{#1}}

\usepackage{hyperref}
\usepackage{url,placeins}

\usepackage[utf8]{inputenc} 
\usepackage[T1]{fontenc}    
\usepackage{hyperref}       
\usepackage{url}            
\usepackage{booktabs,tabularx,multirow}       
\usepackage{amsfonts}       
\usepackage{nicefrac}       
\usepackage{microtype}      
\usepackage{color}
\usepackage{graphicx}
\usepackage{dsfont}
\usepackage{amsmath}

\usepackage{cuted}
\usepackage{capt-of}
\begin{document}

\title{Live Face De-Identification in Video}

%


\maketitle

\begin{abstract}
  We propose a method for face de-identification that enables fully automatic video modification at high frame rates. The goal is to maximally decorrelate the identity, while having the perception (pose, illumination and expression) fixed. We achieve this by a novel feed-forward encoder-decoder network architecture that is conditioned on the high-level representation of a person's facial image. The network is global, in the sense that it does not need to be retrained for a given video or for a given identity, and it creates natural looking image sequences with little distortion in time.
\end{abstract}

\section{Introduction}

In consumer image and video applications, the face has a unique importance that stands out from all other objects. For example, face recognition (detection followed by identification) is perhaps much more widely applicable than any other object recognition (categorization, detection, or instance identification) in consumer images. Similarly, putting aside image processing operators that are applied to the entire frame, face filters remain the most popular filters for consumer video. Since face technology is both useful and impactful, it also raises many ethical concerns. Face recognition can lead to loss of privacy and face replacement technology may be misused to create misleading videos. 
\begin{figure}[t]
  \centering
\includegraphics[width=\linewidth]{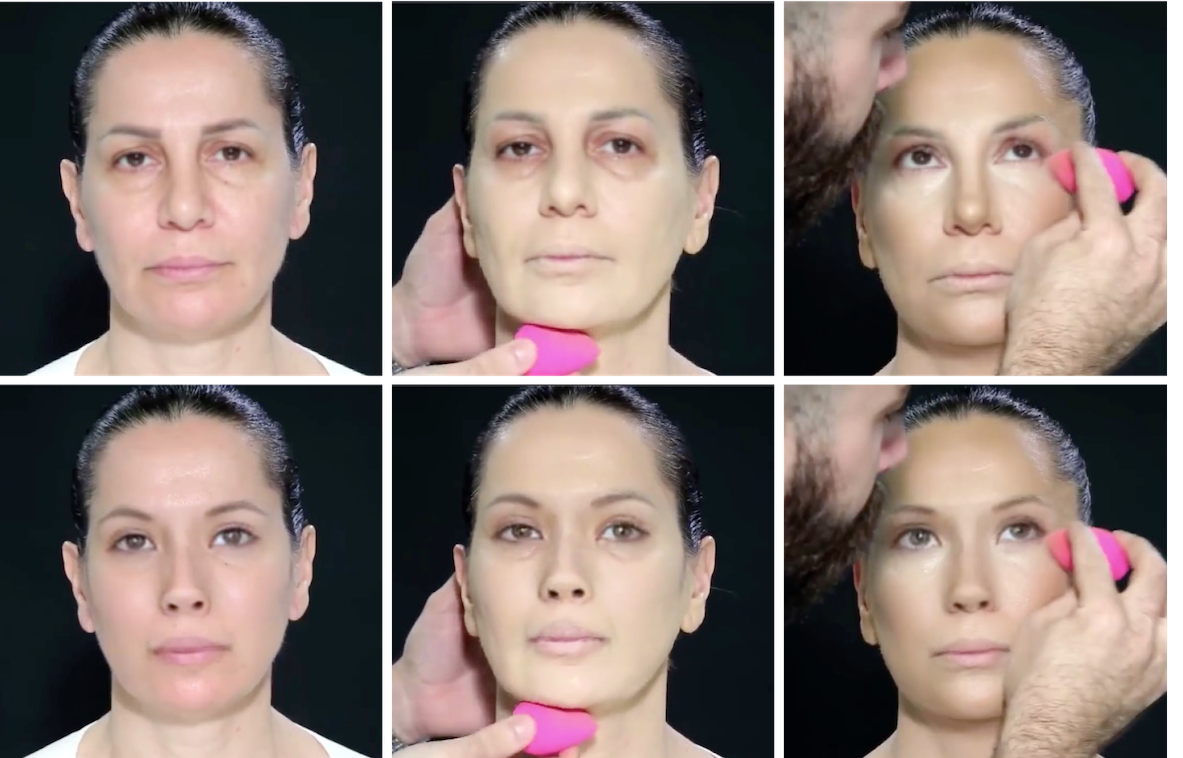}
\captionof{figure}{De-identification video results demonstrated on a variety of {\color{black}poses, expressions}, illumination conditions and occlusions. Pairs of the source frame (first row) and the output frame (second row) are shown. The high-level features (e.g. nose, eyes, eyebrows and mouth) are altered, while the pose, {\color{black}expression, \color{black}lip articulation}, illumination, and skin tone are preserved.
\label{fig:teaser}}
\end{figure}

\begin{table*}[t]
\centering
\begin{small}
\begin{tabular}{l@{~}x{1.4cm}@{~}x{1.3cm}@{~}x{2.0cm}@{~}x{1.9cm}@{~}x{1.6cm}@{~}x{1.2cm}@{~}x{1.4cm}@{~}x{0.9cm}} 
\toprule
&Newton, '05~\cite{1377174}
&Gross, '08~\cite{4587369}
&Samarzija, '14~\cite{6859758}
&Jourabloo, '15~\cite{jourabloo2015attribute}
&Meden, '17~\cite{meden2017face}
&Wu, '18~\cite{ppgan}
&Sun'18 \cite{sun2018natural,sun2018hybrid}
&Our
\\ 
\midrule
Preserves expression            &         -  & - & - & - & - & - & - & + \\
Preserves pose                  &         -  & + & + & - & + & - & + & +\\
Generates new faces     &         -  & $^\dagger$ & - & $^\dagger$ & + & + & + & +\\
Demonstrated on video           &         -  & - & - & - & - & - & - & +\\
Demonstrated on a diverse dataset &	     - & + & - & + & - & - & - & +\\
~~(gender, ethnicity, age, etc.) \\
Reference to a comparison with ours   &  &  & Fig.~\ref{fig:previousfaces} &  & Fig.~\ref{fig:gmms} & Fig.~\ref{fig:ppgan} & Fig.~\ref{fig:hybrid},~\ref{fig:hybrid_supp} & \\
	      
\bottomrule
\end{tabular}
\end{small}
\caption{A comparison to the literature methods. The final row references comparison figures in this work. We compare to all methods that provide reasonable quality images in their manuscript, under conditions that are favorable to previous work (we crop the input images from the pdf files, except for the images received from the authors of~\cite{sun2018natural,sun2018hybrid}). $^\dagger$The face is swapped with an average of a few dataset faces.}
\label{tab:comparisontootherdeid}
\end{table*}

In this work, we focus on video de-identification, which is a video filtering application that both requires a technological leap over the current state-of-the-art, and is benign in nature. This application requires the creation of a video of a similar looking person, such that the perceived identity is changed. This allows, for example, the user to leave a natural-looking video message in a public forum in an anonymous way, that would presumably prevent face recognition technology from recognizing them.

Video de-identification is a challenging task. The video needs to be modified in a seamless way, without causing flickering or other visual artifacts and distortions, such that the identity is changed, while all other factors remain identical, see Fig.~\ref{fig:teaser}. These factors include pose, expression, lip positioning (for unaltered speech), occlusion, illumination and shadow, and their dynamics.

In contrast to the literature methods, which are limited to still images and often swap a given face with a dataset face, our method handles video and generates de novo faces. Our experiments show convincing performance for unconstrained videos, producing natural looking videos. The person in the rendered video has a similar appearance to the person in the original video. However, a state-of-the-art face-recognition network fails to identify the person. A similar experiment shows that humans cannot identify the generated face, even without time constraints.

Our results would not have been possible, without a host of novelties. We introduce a novel encoder-decoder architecture, in which we concatenate to the latent space the activations of the representation layer of a network trained to perform face recognition. As far as we know, this is the first time that a representation from an existing classifier network is used to augment an autoencoder, which enables the feed-forward treatment of new persons, unseen during training. In addition, this is the first work to introduce a new type of attractor-repeller perceptual loss term. This term distinguishes between low- and mid-level perceptual terms, and high-level ones. The former are used to tie the output frame to the input video frame, while the latter is used to distance the identity. In this novel architecture, the injection of the representation to the latent space enables the network to create an output that adheres to this complex criterion. Another unique feature is that the network outputs both an image and a mask, which are used, in tandem, to reconstruct the output frame. The method is trained with a specific data augmentation technique that encourages the mapping to be semantic. Additional terms include reconstruction losses, edge losses, and an adversarial loss.  

\section{Previous Work}

Faces have been modeled by computer graphics systems for a long time. In machine learning, faces have been one of the key benchmarks for GAN-based generative models~\cite{gan,dcgan,gantricks} since their inception. High resolution natural looking faces were recently generated by training both the generator and the discriminator of the GAN progressively, starting with shallower networks and lower resolutions, and enlarging them gradually~\cite{karras2017progressive}.

\begin{figure*}
  \centering
\begin{tabular}{cc}
    \includegraphics[width=0.76\textwidth]{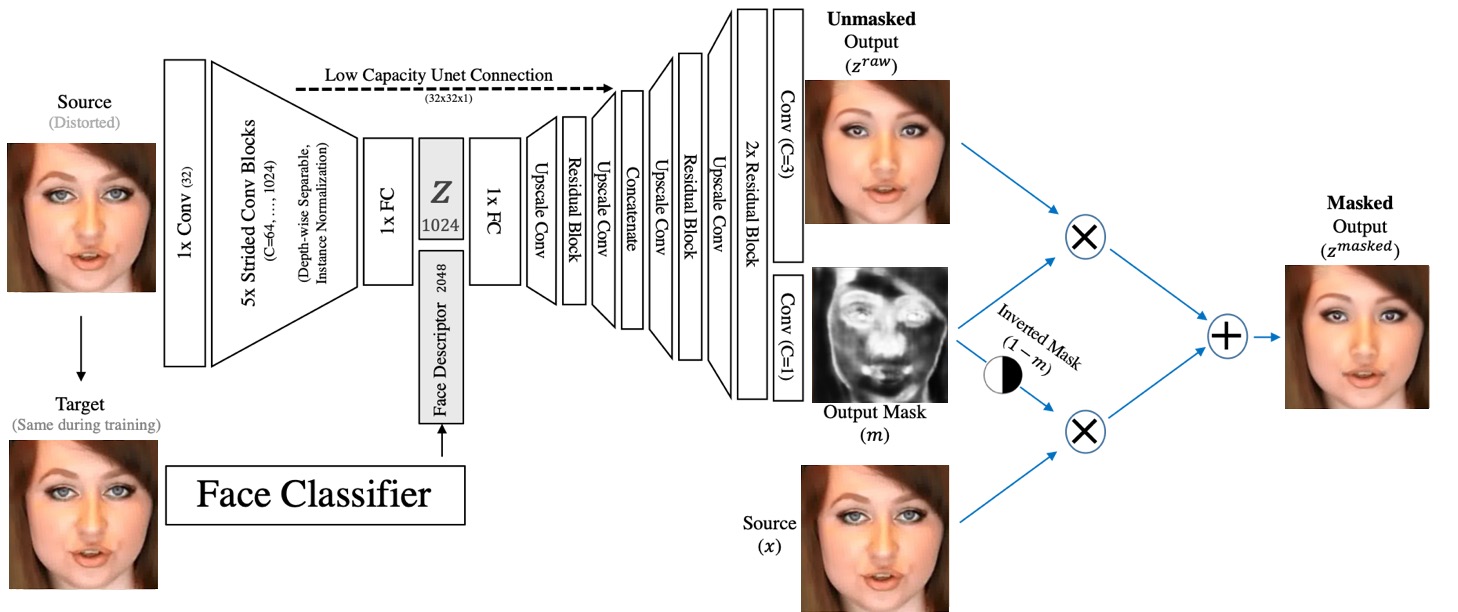}  & 
 \includegraphics[width=0.2\textwidth]{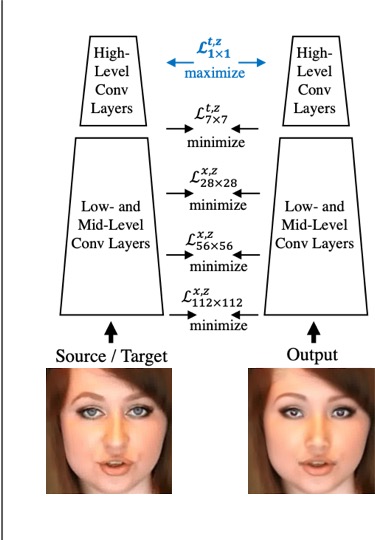} 
\\
    (a) & (b) \\
\end{tabular}
  \caption{(a) The architecture of our network. For conditioning, a pre-trained face recognition network is used. (b) An illustration of the multi-image perceptual loss used, which employs two replicas of the same face recognition network.}
  \label{fig:arch}
\end{figure*}

Conditional generation of faces has been a key task in various unsupervised domain translation contributions, where the task is to learn to map, e.g., a person without eyewear to a person with eyeglasses, without seeing matching samples from the two domains~\cite{discogan,dualgan,distgan,unit}. For more distant domain mapping, such as mapping between a face image and the matching computer graphics avatar, additional supervision in the form of a face descriptor network was used~\cite{02200}. 
Our work uses these face descriptors, in order to distance the identity of the output from that of the input.

As far as we know, our work is the first de-identification work to present results on videos. In still images, several methods have been previously suggested. Earlier work implemented different types of image distortions for face de-identification~\cite{newton2005preserving,4587369}, while more recent works rely on techniques for selecting distant faces~\cite{6859758} or averaging/fusing faces from pre-existing datasets~\cite{1377174,jourabloo2015attribute,meden2017face}. The experiments conducted by the aforementioned techniques are restricted, in most cases, to low-resolution, black and white results. Although it is possible to create eye-pleasing results, they are not robust to different poses, illuminations and facial structures, making them inadequate for video generation. The use of GANs for face de-identification has been suggested~\cite{ppgan}. However, the experiments were restricted to a homogeneous dataset, with no apparent expression preservation within the results. {\color{black}In the GAN-based methods of ~\cite{sun2018natural,sun2018hybrid}, face de-identification is employed for the related task of person obfuscation. The work of~\cite{sun2018natural} conditions the output image based on both a blurred version of the input and the extracted facial pose information. The follow-up work~\cite{sun2018hybrid} combines the GAN-based reconstruction with a parametric face generation network. As both methods are applied over full upper-body images, they result in low facial resolution outputs of $64\times64$. These methods do not preserve expressions, are unsuitable for video, and occasionally provide unnatural outputs.}

Tab.~\ref{tab:comparisontootherdeid} provides a comparative view of the literature. The current literature on de-identification often involves face swapping (our method does not). Face swapping, i.e., the replacement of a person's face in an image with another person's face, has been an active research topic for some time, starting with the influential work of~\cite{blanz2004exchanging,Bitouk:2008:FSA:1399504.1360638}. Recent contributions have shown a great deal of robustness to the source image, as well as for the properties of the image, from which the target face is taken~\cite{Kemelmacher-Shlizerman:2016:TP:2897824.2925871,nirkin2017face}. While these classical face swapping methods work in the pixel space and copy the expression of the target image, recent deep-learning based work swaps the identity, while maintaining the other aspects of the source image~\cite{korshunova2017fast}. In comparison to our work, ~\cite{korshunova2017fast} requires training a new network for every target person, the transferred expression does not show subtleties (which would be critical, e.g., for a speaking person), and the results are not as natural as ours. These limitations are probably a result of capturing the appearance of the target, by restricting the output to be similar, patch by patch, to a collection of patches from the target person. Moreover,~\cite{korshunova2017fast} is limited to stills and was not demonstrated on video. 

The face swapping (FS) project~\cite{dfgithub} is an unpublished work that replaces faces in video in a way that can be very convincing, given suitable inputs. 
Unlike our network, the FS is retrained for every pair of source-video and target-video persons. The inputs to the FS system, during training, are two large sets of images, one from each identity. In order to obtain good results, thousands of images from each individual with a significant variability in pose, expression, and illumination are typically used. In many cases, a large subset of the images of the source person are taken from the video that is going to be converted. In addition, FS often fails, and in order to obtain a convincing output, the person in the source video and the target person need to have a similar facial structure. These limitations make it unsuitable for de-identification purposes.

Like ours, the FS method is based on an encoder-decoder architecture, where both an image and output mask are produced. A few technical novelties of FS are shared with our work. Most notable is the way in which augmentation is performed in order to train a more semantic encoder-decoder network. During the training of FS, the input image is modified by rotating or scaling it, before it is fed to the encoder. The image that the decoder outputs is compared to the undistorted image. 
Another common property is that the GAN variant used employs virtual examples created using the mixup technique~\cite{mixup}. In addition, in order to maintain the pose and expression, which are considered low- or mid-level features in face descriptors (orthogonal to the identity) FS employs a perceptual loss~\cite{perceptual,ulyanov2016texture} that is based on the layers of a face-recognition network.

Another line of work that manipulates faces in video is face reanimation, e.g.,~\cite{thies2016face2face}. This line of work reanimates the face in the target video, as controlled by the face in a source video. This does not provide a de-identification solution in the sense that we discuss -- the output video is reanimated in a different scene, and not in the scene of the source video. In addition, it always provides the same output identity.

We do not enforce disentanglement~\cite{munit,drit,infogan} between the latent representation vector Z and the identity, since the network receives the full information regarding the identity using the face descriptor. Therefore, washing out the identity information in Z may not be beneficial. Similarly, the U-Net connection means that identity information can bypass Z. In our method, the removal of identity is not done through disentanglement but via the perceptual loss. As Fig.~\ref{fig:iflambda} demonstrates, this loss provides a direct and quantifiable means for controlling the amount of identity information. With disentanglement, this effect would be brittle and sensitive to hyperparameters, as is evident in work where the encoding is set to be orthogonal, even to simple multiclass label information, e.g,~\cite{fader}. 

\section{Method}

Our architecture is based on an adversarial autoencoder~\cite{makhzani2015adversarial}, coupled with a trained face-classifier. By concatenating the autoencoder's latent space with the face-classifier representation layer, we achieve a rich latent space, embedding both identity and expression information. The network is trained in a counter-factual way, i.e., the output differs from the input in key aspects, as dictated by the conditioning. The generation task is, therefore, highly semantic, and the loss required to capture its success cannot be a conventional reconstruction loss. 

For the task of de-identification, we employ a target image, which is any image of the person in the video. The method then distances the face descriptors of the output video from those of the target image. The target image does not need to be based on a frame from the input video. This contributes to the applicability of the method, allowing it to be applied to live videos. In our experiments, we do not use an input frame in order to show the generality of the approach. 
To encode the target image, we use a pre-trained face classifier ResNet-50 network~\cite{resnet}, trained over the VGGFace2 dataset~\cite{vggface2}.

The process during test time is similar to the steps taken in the face swapping literature and involves the following steps:
(a) A square bounding box is extracted using the 'dlib'~\cite{dlib} face detector. 
(b) 68 facial points are detected using~\cite{kazemi2014one}. (c) A transformation matrix is extracted, using an estimated similarity transformation (scale, rotation and translation) to an averaged face. (d) The estimated transformation is applied to the input face. (e) The transformed face is passed to our network, together with the representation of the target image, obtaining both an output image and a mask. (f) The output image and mask are projected back, using the inverse of the similarity transformation. (g) We generate an output frame by linearly mixing, per pixel, the input and the network's transformed output image, according to the weights of the transformed mask. (h) The outcome is merged into the original frame, in the region defined by the convex hull of the facial points.

At training time, we perform the following steps: (a)  The face image is distorted and augmented. This is done by applying random scaling, rotation and elastic  deformation. 
(b) The distorted image is fed into the network, together with the representation of a target image. During training, we select the same image, undistorted. 
(c) A linear combination of the masked output (computed as in step (g) above) and the undistorted input is fed to the discriminator. This is the mixup technique~\cite{mixup} discussed below.
(d) Losses are applied on the network's mask and image output, as well as to the masked output, as detailed below.

Note that there is a discrepancy between how the network is trained and how it is applied. Not only do we not make any explicit effort to train on videos, the target images are selected in a different way. 
During training, we extract the identity from the training image itself and not from an independent target image. The method is still able to generalize to perform the real task on unconstrained videos.

\subsection{Network architecture}

The architecture is illustrated in Fig.~\ref{fig:arch}(a). The encoder is composed of a convolutional layer, followed by five strided, depth-wise separable~\cite{xception} convolutions with instance normalization~\cite{ulyanov2016instance}. Subsequently, a single fully connected layer is employed, and the target face representation is concatenated. 
The decoder is composed of a fully connected layer, followed by a lattice of upscale and residual~\cite{resnet} blocks, terminated with a $tanh$ activated convolution for the output image, and a sigmoid activated convolution for the mask output. Each upscale block is comprised of a 2D convolution, with twice the number of filters as the input channel size. Following an instance normalization and a LReLU~\cite{delving} activation, the activations are re-ordered, so that the width and height are doubled, while the channel size is halved. Each residual block input is summed with the output of a Conv2D-LReLU-Conv2D chain.

A low-capacity U-net connection~\cite{unet} is employed (32x32x1), thus relieving the autoencoder's bottleneck, allowing a stronger focus on the encoding of transfer-related information. The connection size does not exceed the bottleneck size (1024) and due to the distortion of the input image, a collapse into a simple reconstructing autoencoder in early training stages is averted. 

The discriminator consists of four strided convolutions with LReLU activations, with instance normalization applied on all but the first one. A sigmoid activated convolution yields a single output.

The network has two versions: a lower resolution version generating 128x128 images, and a higher resolution version, generating 256x256 images. The higher resolution decoder is simplified and enlarged and consists of a lattice of 6x(Upscale block --> Residual block). {\color{black} Unless otherwise specified, the results presented in the experiments are done with the high-res model.}

\subsection{Training and the Losses Used}

For training all networks, except for the discriminator $D$, we use a compound loss $\mathcal L$, which is a weighted sum of multiple parts:
\begin{align*}
\nonumber
\mathcal L = &\alpha_0\mathcal L_G + \alpha_1 \mathcal L^{raw}_{R} + \alpha_1 \mathcal L^{masked}_{R} + \alpha_2 \mathcal{L}_{x}^{raw}\\ & + \alpha_2 \mathcal{L}_{y}^{raw} + \alpha_2 \mathcal{L}_{x}^{masked} + \alpha_2 \mathcal{L}_{y}^{masked}  \nonumber \\  & +\alpha_3 \mathcal L_p^{raw} +  \alpha_3 \mathcal L_p^{masked} + \alpha_4   \mathcal{L}^m + \alpha_5 \mathcal{L}^m_{x} + \alpha_5 \mathcal{L}^m_{y},
\end{align*}

where $\mathcal L_G$ is the generator's loss, $\mathcal L_{R}^{raw}$ and $\mathcal L_{R}^{masked}$ are reconstruction losses for the output image of the decoder $z^{raw}$ and the version after applying the masking $z^{masked}$, $\mathcal{L}^{*}_{x}$ and $\mathcal{L}^{*}_{y}$ are reconstruction losses applied to the spatial images derivatives, $\mathcal L_p^{*}$ are the perceptual losses, and $\mathcal L^m_{*}$ are regularization losses on the mask. The discriminator network is trained using its own loss $\mathcal L_D$. Throughout our experiments, we employ $\alpha_0=\alpha_1=\alpha_2=\alpha_3=0.5,\alpha_4=3\cdot10^{-3}, \alpha_5=10^{-2}$.
 
To maintain realistic looking generator outputs, an  adversarial loss is used with a convex combination of example pairs (known as mixup)~\cite{mixup} over a Least Square GAN~\cite{lsgan}: 
\begin{align*}
  \mathcal{L}_{D}  &=\|D(\delta_{mx}) - \lambda_{\beta} \mathds{1}\|_2^2 \\
  \mathcal{L}_{G}  &=\alpha_0 \|D(\delta_{mx}) - (1 - \lambda_{\beta}) \mathds{1}\|_2^2  \end{align*}
While, $\delta_{mx} = \lambda_{\beta} \cdot x + (1 - \lambda_{\beta}) z^{masked}$ and $\lambda_\beta$ is sampled out of a Beta distribution $\lambda_{\beta} \sim Beta (\alpha,\alpha)$, $x$ is the undistorted input ``real'' sample and $z^{masked}$ is the post masking generated sample. A value of $\alpha=0.2$ is used throughout the experiments.

Additional losses are exercised to both retain source-to-output similarity, yet drive a perceptible transformation. Several losses are distributed equally between the raw and masked outputs, imposing constraints on both. 
An L1 reconstruction loss is used to enforce pixel-level similarity:
\begin{align*}
\mathcal{L}_{R}^{raw}
=\alpha_1 \|z^{raw} - x\|_1 & & 
\mathcal{L}_{R}^{masked}
=\alpha_1 \|z^{masked} - x \|_1 
\end{align*}
where $z^{raw}$ is the output image itself. 
This results in a non-trivial constraint, as the encoder input image is distorted. An edge-preserving loss is used to constrain pixel-level derivative differences in both the $x$ and $y$ image axes. Calculated as the absolute difference between the source and output derivatives in each axis direction for both the raw and masked outputs:
\begin{align*}
\mathcal{L}_{x}^{raw}
=\alpha_2 \|z^{raw}_{x} - x_{x}\|_1 & & 
\mathcal{L}_{x}^{masked}
=\alpha_2 \|z^{masked}_{x} - x_{x}\|_1 \nonumber\\
\mathcal{L}_{y}^{raw}=\alpha_2 \|z^{raw}_{y} - x_{y}\|_1 & & 
\mathcal{L}_{y}^{masked}
=\alpha_2 \|z^{masked}_{y} - x_{y}\|_1 
\end{align*}
where $x_x$ is the derivative of the undistorted input image $x$ along the $x$ axis, and similarly for outputs $z$ and the $y$ axis.

Additional losses are applied to the blending mask $m$, where $0$ indicates that the value of this pixel would be taken from the input image $x$, $1$ indicates taking the value from $z^{raw}$, and intermediate values indicate linear mixing. We would like the mask to be both minimal and smooth and, therefore, employ the following losses:
\begin{align*}
\mathcal{L}^m=\|m\|_1 & & 
\mathcal{L}^m_x= \|m_{x}\|_1 & & 
\mathcal{L}^m_y= \|m_{y}\|_1
\end{align*}
where $m_x$ and $m_y$ are the spatial derivatives of the mask.

\subsubsection{A Multi-Image Perceptual Loss}

A new variant of the perceptual loss~\cite{perceptual} is employed to maintain source expression, pose and lighting conditions, while capturing the target identity essence. This is achieved by employing a perceptual loss between the undistorted source and generated output on several low-to-medium abstraction layers, while distancing the high abstraction layer perceptual loss between the target and generated output.

Let $ a^r_{n\times n} $ be the activations of an $ n\times n $ spatial block within the face classifier network for image $r$, where in our case, $r$ can be either the input image $x$, the target image $t$, the raw output $z^{raw}$, or the masked output $z^{masked}$.  

We consider the spatial activations maps of size $112\times 112, 56\times 56, 28\times 28$ and $7\times 7$, as well as the representation layer of size $1\times 1$. The lower layers (larger maps) are used to enforce  similarity to the input image $x$, while the $7 \times 7$ layer is used to enforce similarity to $t$, and the $1\times 1$ feature vector is used to enforce dissimilarity to the target image. 

Let us define $\ell^{r_1,r_2}_{n \times n} = c_{n} \| a_{r_1, n\times n}- a_{r_2, n\times n}\|_1$, where $c_n$ is a normalizing constant, corresponding to the size of the spatial activation map.

The perceptual loss is given by:
\begin{align*}
\mathcal L^c_p &= \ell^{x,z^{c}}_{112\times 112} + \ell^{x,z^{c}}_{56\times 56} + \ell^{x,z^{c}}_{28\times 28} + \ell^{t,z^{c}}_{7\times 7} - \lambda \ell^{t,z}_{1\times 1} 
\end{align*}
for $c$ that is either $raw$ or $masked$, and where $\lambda>0$ is a hyperparameter, which determines the generated face's high level features distance from those of the target image.

The application of the multi-image perceptual loss during training is depicted in Fig.~\ref{fig:arch}(b). During training, the target is the source, and there is only one input image. The resulting image has the texture, pose and expression of the source, but the face is modified to distance the identity. 
Note that we refer to it as a multi-image perceptual loss, as its aim is to minimize the analog error term during inference (generalization error). However, as a training loss, it is only applied during train, where it receives a pair of images, similar to other perceptual losses. 

\begin{figure}[t]
  \centering
 \includegraphics[width=.90124\linewidth]{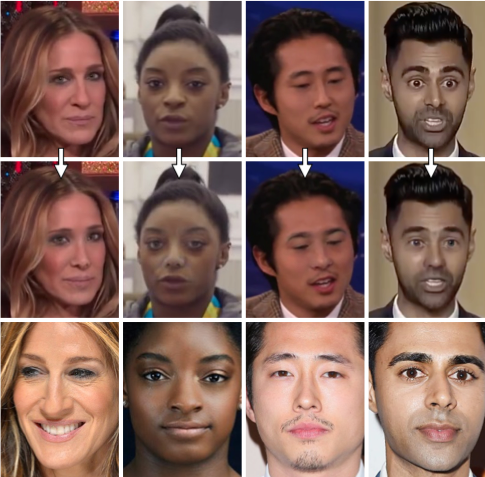} 
  \caption{Sample results for video de-identification (zoom). Triplets of source frame, converted frame and target are shown. The modified frame looks similar but the identity is completely different.}
  \label{fig:plasticresults}
  \vspace{-.5cm}
 \end{figure}

{\color{black}Note that the perceptual loss parameters $c_n$ are normalizing constants obtained by counting the number of elements. In addition, $\alpha_0=\alpha_1=\alpha_2=\alpha_3$ are simply set to one, and $\alpha_4,\alpha_5$ were chosen arbitrarily. Therefore, there is effectively, only a single important hyperparameter: $\lambda$, which provides a direct control of the strength of the identity distance which requires tuning (see Fig.~\ref{fig:iflambda}).}

At inference time, the network is fed an input frame and a target image. The target image is transmitted through the face classifier, resulting in a target feature vector, which, in turn, is concatenated to the latent embedding space. 
Due to the way the network is trained, the decoder will drive the output image away from the target feature vector.

\section{Experiments}
\begin{table*}[t]
\begin{tabularx}{\textwidth}{c@{~~~}c@{~}c@{~}c}
\begin{small}
\begin{tabular}[b]{@{}lcc@{}}
    \toprule
    Video     & lower & higher \\
    \midrule
1&	28.7\% & 34.2\% \\
2&	66.7\% & 45.8\%\\
3&	61.9\% & 64.3\%\\
4&	52.4\% & 62.1\%\\
5&	42.9\% & 43.8\%\\
6&	47.6\% & 27.0\%\\
7&	57.1\% & 56.8\%\\
8&	71.4\% & 73.5\%\\
    \bottomrule
  \end{tabular} 
  \end{small}%
&
\raisebox{.35cm}{\includegraphics[width=.2954\linewidth]{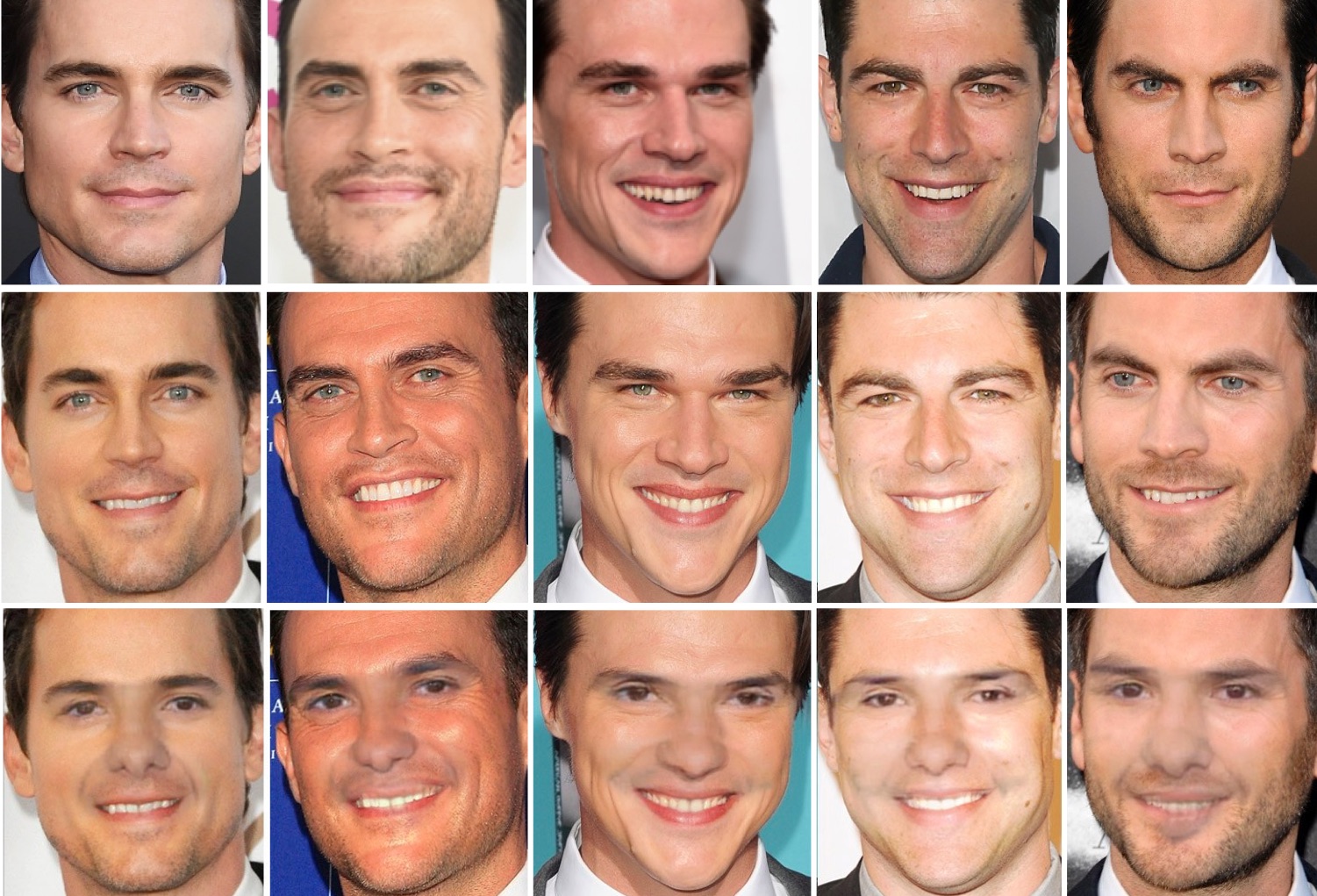}} &
\includegraphics[width=0.223154\textwidth]{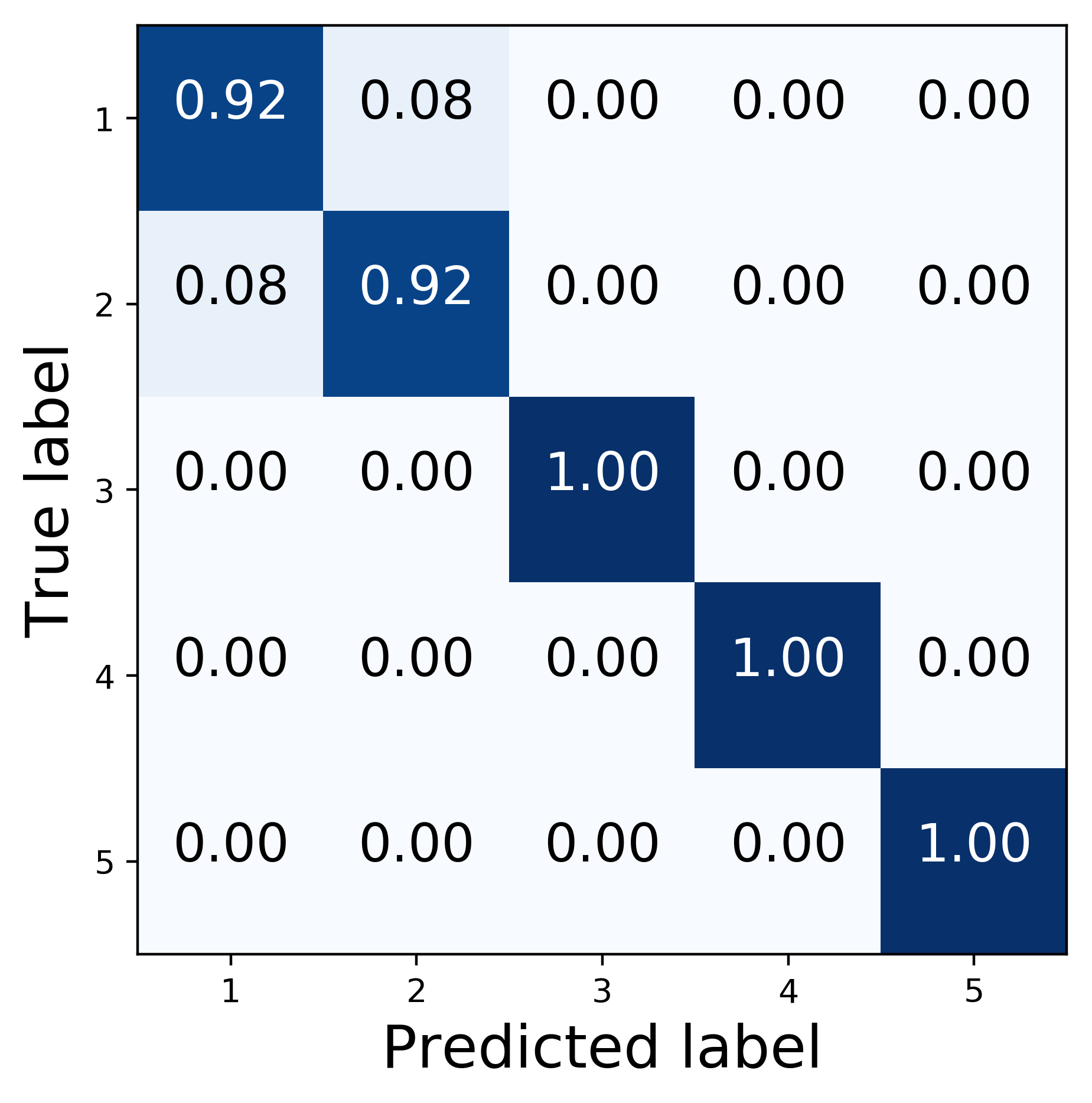}  &
          \includegraphics[width=0.223154\textwidth]{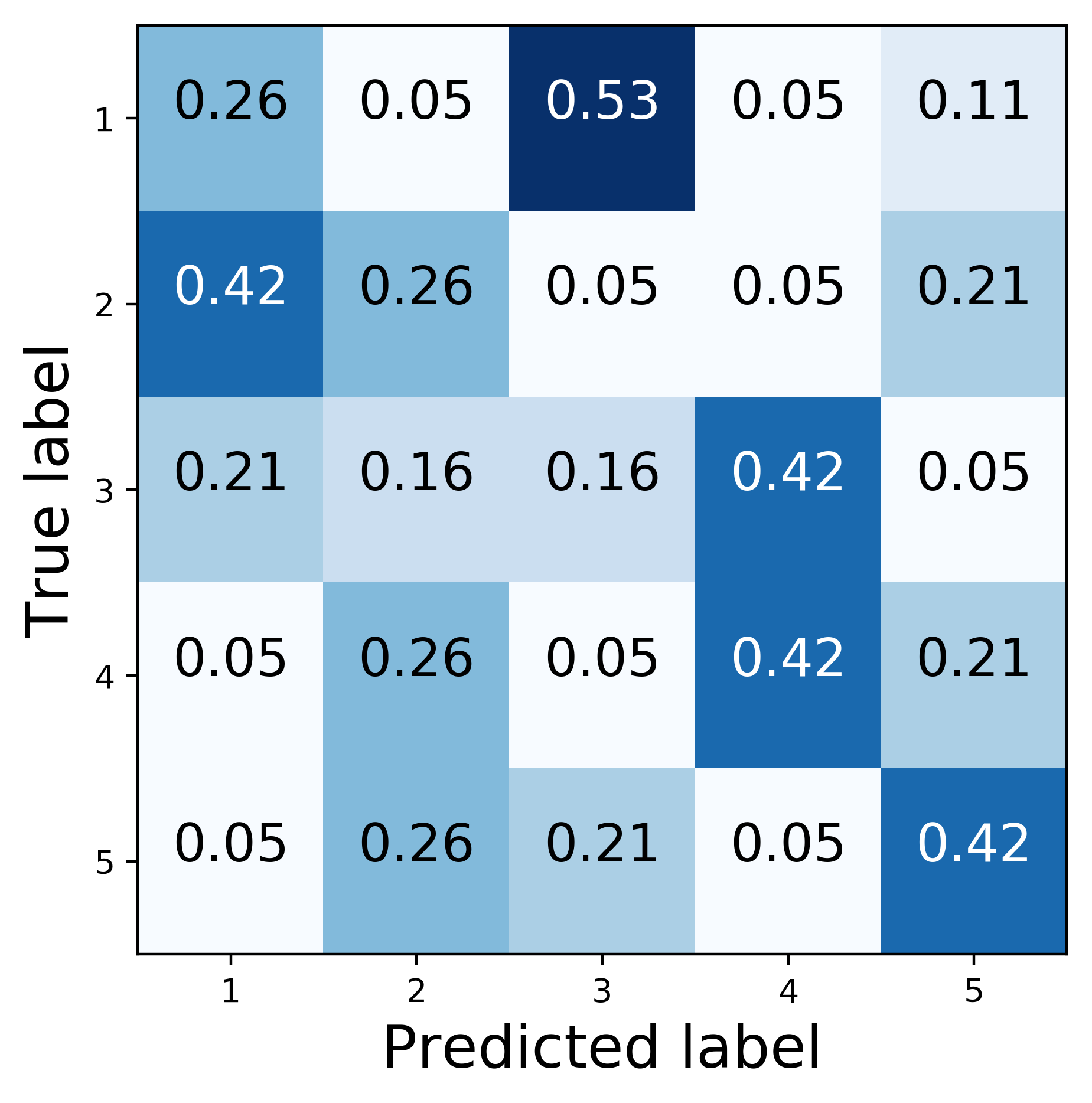} 
 \\
  (a) & (b) & (c)& (d)\\
\end{tabularx}
  \caption{(a) Video user study - Success rate in user identification of a real video from a modified one for both lower-resolution and higher-resolution models. Closer to 50\%  is better. (b) Each column is a different individual from the still image user study. [Row 1] The gallery images, i.e, the album images the users were asked to select the identity from. [Row 2] The input images. [Row 3] The de-identified version of [Row 2]. (c) The confusion matrix in identifying the five persons for the real images (control). (d) The confusion matrix for identifying, based on the de-identified images. }
  \label{tab:user}
\end{table*}
\begin{table}[t]
\begin{center}
\begin{tabular}{llcc}
\toprule
& & \small RGB & \small Face\\
Person in&\small Method  & \small values & \small desc.\\
\midrule
\multirow{ 2}{*}{Row 1} & \small \cite{6859758}       & \small 5.46 & \small 1.21 \\
&\small Our high      & \small 2.72 & \small 1.50 \\
\midrule
\multirow{ 2}{*}{Row 2} & \small \cite{6859758}       & \small 4.91 & \small 1.35 \\
&\small Our high      & \small 2.35 & \small 1.53 \\
\midrule
\multirow{ 2}{*}{Row 3} & \small \cite{6859758}       & \small 4.51 & \small 1.20 \\
&\small Our high      & \small 3.92 & \small 1.32 \\
\bottomrule
\end{tabular}
\end{center}
\caption{The distance between the original and de-identified image, for the images in Fig.~\ref{fig:previousfaces}. Our method results in lower pixel differences but with face descriptor distances that are higher.}
 \label{tab:previousfaces}
\end{table}
\begin{table*}[t]
\begin{center}
  \begin{tabular}{lcccccc}
    \toprule
    & \multicolumn{2}{c}{Original frames} & \multicolumn{2}{c}{Lower-res de-ID model}      &  \multicolumn{2}{c@{}}{Higher-res de-ID }\\
    \cmidrule(lr){2-3}
    \cmidrule(lr){4-5}
    \cmidrule(lr){6-7}
    Person     & Median  & Mean    &  Median  & Mean    & Median & Mean     \\
               &  & $\pm$SD &   & $\pm$SD &  & $\pm$SD \\
    \midrule
    Simone Biles & 1 & 3 $\pm{50}$ & 1730 &  2400.6$\pm{2142}$ & 1725 & 2223$\pm{1814}$  \\
    Billy Corgan & 1  & 95.6$\pm{313}$  &3156 & 3456.3$\pm{2601}$& 901 & 1334$\pm{1518}$\\
    Selena Gomez & 1 & 1 $\pm{0}$ &  2256 & 2704$\pm{1873}$ & 8058 & 8110$\pm{2186}$ \\
    Scarlett Johansson & 1 &  3.8$\pm{38.6}$ & 9012 &  7753.5$\pm{3112}$ & 4493 & 4830$\pm{2544}$ \\
    Steven Yeun & 1  & 1.02$\pm{0.6}$  & 5806 & 4976.2$\pm{3167}$ & 1069 & 1814$\pm{2544}$ \\
    Sarah J. Parker & 1 & 1$\pm{0}$  &  679 & 1069.3$\pm{1096}$ & 408 & 620$\pm{665}$\\
    \midrule
    Average    & 1 & 17 & 3773 & 3726 & 2776 & 3155 \\
    \bottomrule
  \end{tabular}
  \end{center}
  \caption{Ranking of the true identity out of a dataset of 54,000	persons (SD=Standard Deviation). Evaluation is performed on the pre-trained LResNet50E-IR ArcFace network. Results are given for both the lower- and higher-resolution models.}
  \label{tab:ranking}
\end{table*}

Training is performed using the Adam~\cite{adam} optimizer, with the learning rate set to $10^{-4}$, $\beta_1=0.5$, and $\beta_2=0.99$. At each training iteration, a batch of 32 images for the lower resolution model, and 64 for the higher resolution model, are randomly selected and augmented. We initialize all convolutional weights using a random normal distribution, with a mean of 0 and a standard deviation of 0.02. Bias weights are not used. The decoder includes LReLU activations with $\alpha=0.2$ for residual blocks and $\alpha=0.1$ otherwise. {\color{black} The low-resolution} network was trained on a union of LFW~\cite{lfw}, CelebA~\cite{celeba} and PubFig~\cite{pubfig}, totaling 260,000 images, the vast majority from CelebA. The identity information is not used during training. {\color{black} The high-resolution network was train on a union of CelebA-HQ~\cite{karras2017progressive}, and faces extracted out of the 1,000 source videos used by~\cite{roessler2018faceforensics}, resulting in 500,000 images}. {Training was more involved for the lower resolution model, and it was trained for 230k iterations with a gradual increasing strength of the hyperparameter $\lambda$, ranging from $\lambda=1\cdot 10^{-7}$ to $\lambda=2\cdot 10^{-6}$, in four steps. Without this gradual increase, the naturalness of the generated face is diminished. For the higher resolution model, 80k iterations with a fixed $\lambda=2\cdot 10^{-6}$ were sufficient.}

Sample results are shown in Fig.~\ref{fig:plasticresults}. In each column, we show the original frame, the modified (output) frame, and the target image from which the identity was extracted. As can be seen, our method produces natural looking images that match the input frame. Identity is indeed modified, while the other aspects of the frame are maintained. 

The supplementary media (\url{https://youtu.be/cCYnBtni7Wg}) 
contains sample videos, with significant motion, pose, expression and illumination changes, to which our method was applied. Our method can deal with videos, without causing motion- or instability-based distortions. This is despite being strictly based on per-frame analysis. 

It is also evident that the lower resolution model seems blurry at times. This is a consequence of the fixed resolution and not of the generated image, which is in fact sharp. The higher resolution model clearly provides more pleasing results, when the required resolution is high.

To test the naturalness of the approach, we tested the ability of humans to discriminate between videos that were modified to those that were not. {\color{black} Although the human observers ($n=20$) were fully aware of the type of manipulation that the videos had undergone,} the human performance {\color{black} was} close to random, with an average success rate of 53.6\% (SD=13.0\%), see Tab.~\ref{tab:user}(a). In order to avoid a decision based on a familiar face, this was evaluated on a non-celebrity dataset created specifically for this purpose, which contained 8 videos.

Familiar identities, can often be recognized by non-facial cues. To establish that given a similar context around a facial identity (e.g. hair, gender, ethnicity), the perceived identity is shifted in a way that is almost impossible to place, we considered images of five persons of the same ethnicity and similar hair styles from a TV show, and collected two sets of images: reference (gallery) and source. The source images were modified by our method, using them as targets as well, see Tab.~\ref{tab:user}(b). As can be seen in the confusion matrix of Tab.~\ref{tab:user}(c), the users could easily identify the correct gallery images, based on the source images. However, as Tab.~\ref{tab:user}(d) indicates, post de-identification, the answers had little correlation with the true identity, as desired.

In order to automatically quantify the performance of our de-identification method, we applied a state-of-the-art face-recognition network, namely, the ArcFace~\cite{arcface} LResNet50E-IR network. This network was selected both for its performance, and for the dissimilarity between this network and the VGGFace2 network, used as part of our network, in both the training set and loss. 

The results of the automatic identification are presented in Tab.~\ref{tab:ranking} for both the lower resolution and the higher resolution models. Identification is performed out of the 54,000 persons in the ArcFace verification set. The table reports the rank of the true person out of all persons, when sorting the softmax probabilities that the face recognition network produces. The ranking of the true identity in the original video shows an excellent recognition capability, with most of the frames identifying the correct person as the top-1 result. For the de-identified frames, despite the large similarity between the original and the modified frames (Fig.~\ref{fig:plasticresults}), the rank is typically in the thousands.

Another automatic face recognition experiment is conducted on the LFW benchmark~\cite{lfw}. Tab.~\ref{tab:lfw_exp} presents the results on de-identified LFW image pairs for a given person (de-identification was applied to the second image of each pair), for two FaceNet~\cite{schroff2015facenet} models. 
The true positive rate for the LFW benchmark drops from almost $0.99$, to less than $0.04$ after applying de-identification. 

\begin{figure}[t]
\begin{tabular}{lcc}
    \toprule
  FaceNet Model       & Original  & De-ID \\
    \midrule
     VGGFace2 & $0.986\pm0.010$ & ${\bf{0.038}}\pm 0.015$\\
     CASIA & $0.965\pm0.016$ & ${\bf{0.035}}\pm0.011$ \\

    \bottomrule
  \end{tabular}
\captionof{table}{Results on the LFW benchmark, employing the FaceNet network trained on VGGFace2 or CASIA-WebFace. Shown is the True Positive Rate for a False Acceptance Rate of $0.001$.}
 \label{tab:lfw_exp}
 \end{figure}
{\color{black}An additional experiment, evaluating our method on the LFW benchmark, can be found in the appendix.}

\begin{figure}[t]

    \begin{minipage}[c]{0.52\linewidth}
    \begin{tabular}{@{}c@{~}c@{~}c} 
  \includegraphics[width=1.445682cm]{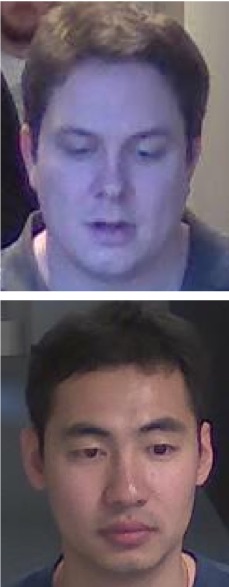} & \includegraphics[width=1.445682cm]{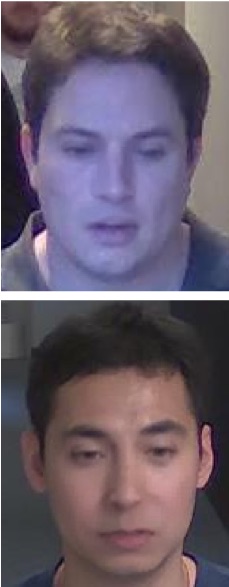} & \includegraphics[width=1.445682cm]{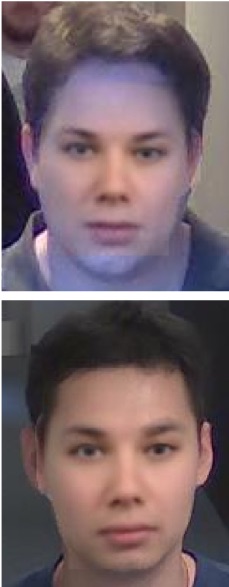} \\
  (a) & (b) & (c)
  \end{tabular}
  \caption{(a) Input images from~\cite{meden2017face}, (b) our results, (c) those of~\cite{meden2017face}. Our method maintains the expression, pose, and illumination. Furthermore, our work does not assign the same new identity to different persons.
  } 
  \label{fig:gmms}
\end{minipage}%
\hfill
\begin{minipage}[c]{0.44\linewidth}
  \begin{tabular}{@{}c@{~}c@{~}c} 
  \includegraphics[height=4.901059cm]{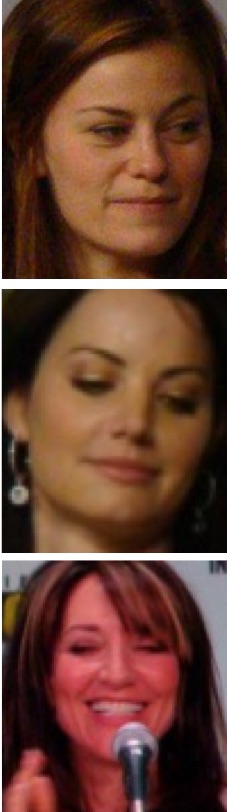} & \includegraphics[height=4.901059cm]{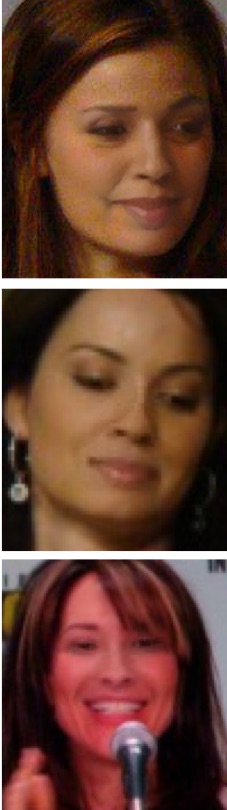} & \includegraphics[height=4.901059cm]{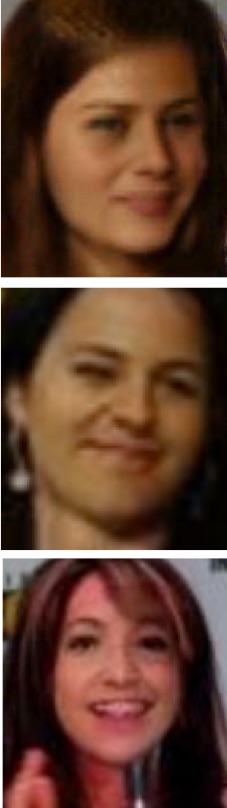} \\
  (a) & (b) & (c)
  \end{tabular}
\caption{(a) Input images from~\cite{sun2018natural,sun2018hybrid}, (b) our results, (c) those of~\cite{sun2018natural} (row 1) and ~\cite{sun2018hybrid} (rows 2-3). }
\label{fig:hybrid}
 \end{minipage}

\end{figure}

{\color{black}A comparison of our method with the recent work of ~\cite{meden2017face} is given in Fig.~\ref{fig:gmms}. This method relies on the generation of a new identity, given the k-closest identities, as selected by a trained CNN feature-extractor. As can be seen, this can result in the same rendered identities for multiple inputs, and does not maintain the expression, illumination and skin tone.}

\begin{figure}[t]
  \centering
\includegraphics[width=.95\linewidth]{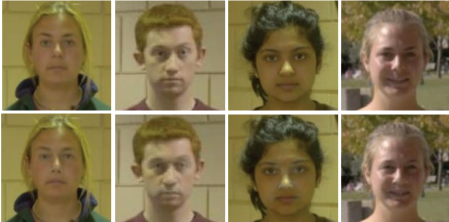}  
\caption{De-Identification applied to the examples labeled as very challenging in the NIST Face Recognition Challenge~\cite{gbu}.}
  \label{fig:gbu}
\centering
\begin{tabular}{c@{~}c@{~}c@{~}c}
    \includegraphics[width=1.92cm]{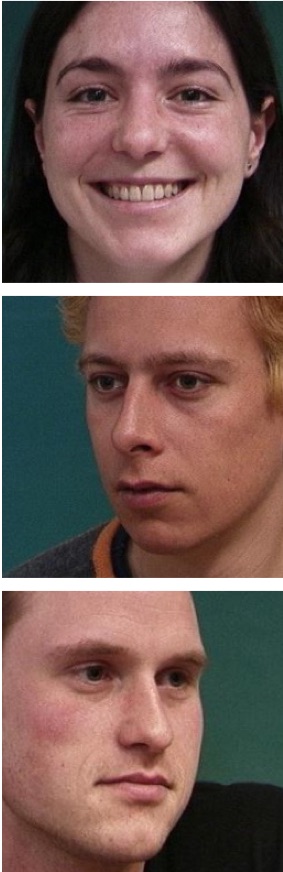} & 
    \includegraphics[width=1.92cm]{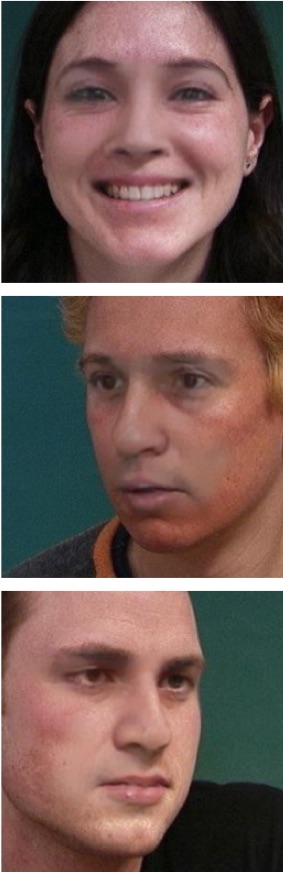} & 
    \includegraphics[width=1.92cm]{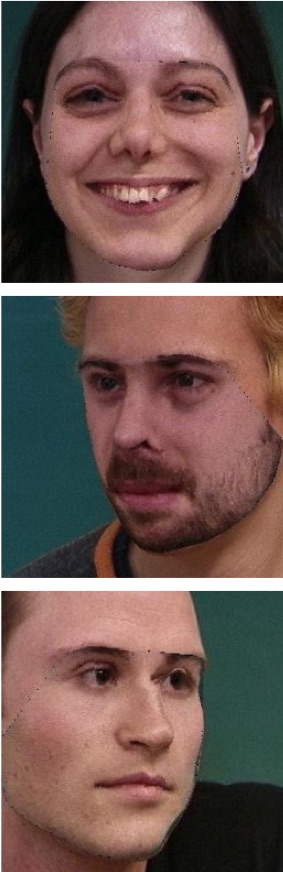} & 
    \includegraphics[width=1.9185cm]{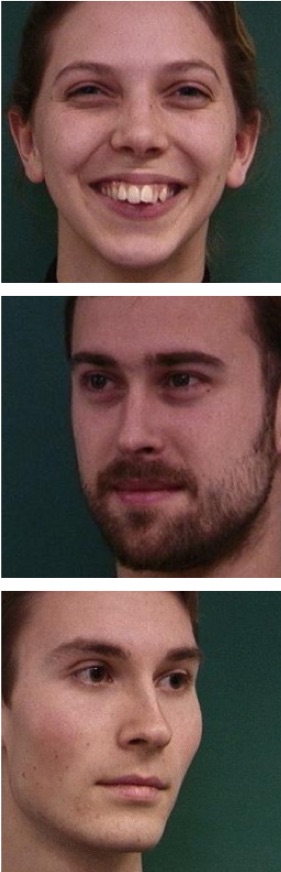}   \\
  (a) & (b) & (c) & (d)
  \end{tabular}
  \caption{Comparison with~\cite{6859758} (from the paper sample image). (a) Original image (also used for the target of our method). (b) Our generated output. (c) Result of ~\cite{6859758}. (d) Target used by~\cite{6859758}.}
  \label{fig:previousfaces}
\centering
\includegraphics[width=.85\linewidth]{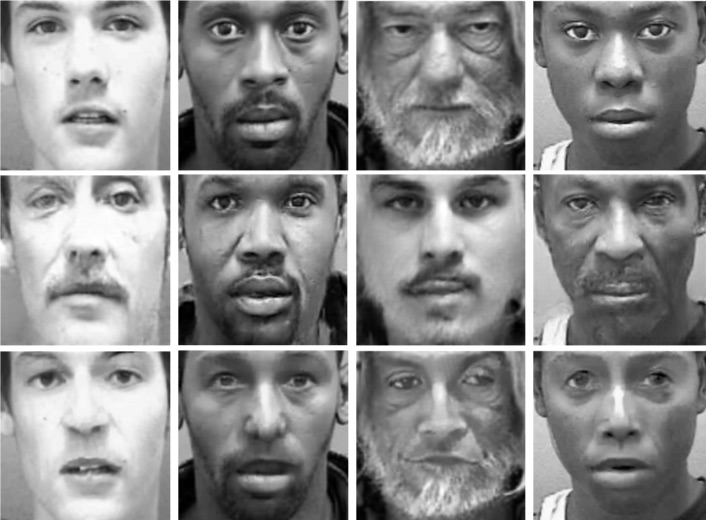} 
\caption{Comparison with~\cite{ppgan}. Row 1 - Original images. Row 2 - results of ~\cite{ppgan}. Row 3 - Our generated outputs. The previous work does not maintain mouth expression or facial hair.}
  \label{fig:ppgan}

  \centering
  \begin{tabular}{@{}c@{~}c@{~}c@{~}c@{~}c@{}}
  \includegraphics[width=.24\linewidth]{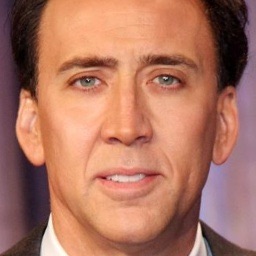} &  \includegraphics[width=.24\linewidth]{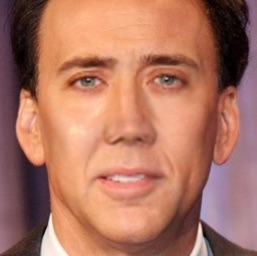} & \includegraphics[width=.24\linewidth]{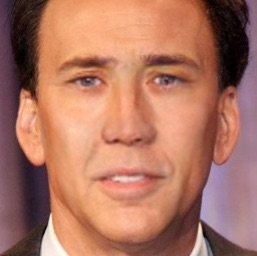} & \includegraphics[width=.24\linewidth]{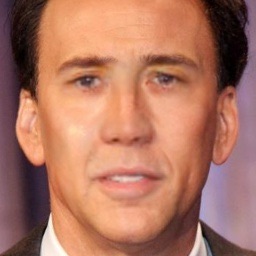}\\ 
  (a) & (b) & (c) & (d)
  \end{tabular}
  \caption{Incrementally growing $\lambda$ in the lower resolution model. A gradual identity shift can be observed. (a) Source. (b) $\lambda=-5\cdot 10^{-7}$. (c) $\lambda=-1\cdot 10^{-6}$. (d) $\lambda=-2\cdot 10^{-6}$.}
  \label{fig:iflambda}
\end{figure}

To emphasize the ability of identity-distancing, while maintaining pixel-space similarity, we compare our method to~\cite{6859758}. While the method of \cite{6859758} relies on finding a dissimilar identity within a given dataset, ours is single-image dependent, in the sense that it does not rely on other images within a dataset. It is, therefore, resilient to different poses, expressions, lighting conditions and face structures. Given the figures provided in the work of \cite{6859758}, we compare our generated outputs by high-level perceptual distance from the source face, taking into account pixel-level similarity (Fig.~\ref{fig:previousfaces}). A comparison of the distance between the original and the de-identified image for the two methods (Tab.~\ref{tab:previousfaces}) reveals that our method results in lower pixel differences, yet with face descriptor distances that are higher. 

A comparison with the work of~\cite{ppgan} is given in Fig.~\ref{fig:ppgan}. Our results are at least as good as the original ones, despite having to run on the cropped faces extracted from the paper PDF. Although ~\cite{ppgan} presents visually pleasing results, they do not maintain low-level and medium-level features, including mouth expression and facial hair. In addition, the work of ~\cite{ppgan} presents results on low-resolution black and white images only, with no pose or gender variation.

Figure \ref{fig:hybrid} compares with the recent work of~\cite{sun2018natural,sun2018hybrid}. Our method is able to distance the identity in a more subtle way, while introducing less artifact. Our generated image contains only the face, which is enabled by the use of the mask. Their method generates both the face and the upper body using the same $256\times 256$ generation resolution, which makes our results of a much higher effective resolution. {\color{black}A full set of results is given in the {\color{black}appendix, Fig.~\ref{fig:hybrid_supp}}.}

To further demonstrate the robustness of our method, we applied our technique to images copied directly from the very difficult inputs of ~\cite{gbu}. As can be seen in Fig.~\ref{fig:gbu}, our method is robust to very challenging illuminations. 

To demonstrate the control of the hyperparameter $\lambda$ over the identity distance, we provide a sequence of generated images, where each trained model is identical, apart from the strength of $\lambda$. The incremental shift in identity can be seen in Fig.~\ref{fig:iflambda}. {\color{black} Ablation analyses are given in the appendix. The analyses compare various variants of our method, and depict the artifacts introduced by removing parts of it.}


\section{Conclusions}

Recent world events concerning the advances in, and abuse of face recognition technology invoke the need to understand methods that successfully deal with de-identification. Our contribution is the only one suitable for video, including live video, and presents quality that far surpasses the literature methods. The approach is both elegant and markedly novel, employing an existing face descriptor concatenated to the embedding space, a learned mask for blending, a new type of perceptual loss for getting the desired effect, among a few other contributions.

Minimally changing the image is important for the method to be video-capable, and is also an important factor in the creation of adversarial examples~\cite{oh2017adversarial}. Unlike adversarial examples, in our work, this change is measured using low- and mid-level features and not using norms on the pixels themselves. It was recently shown that image perturbations caused by adversarial examples distort mid-level features~\cite{xie2018feature}, which we constrain to remain unchanged.

\clearpage

{\small
\bibliographystyle{ieee_fullname}
\bibliography{main}

\begin{thebibliography}{10}\itemsep=-1pt

\bibitem{distgan}
Sagie Benaim and Lior Wolf.
\newblock One-sided unsupervised domain mapping.
\newblock In {\em NIPS}, 2017.

\bibitem{Bitouk:2008:FSA:1399504.1360638}
Dmitri Bitouk, Neeraj Kumar, Samreen Dhillon, Peter Belhumeur, and Shree~K.
  Nayar.
\newblock Face swapping: Automatically replacing faces in photographs.
\newblock In {\em SIGGRAPH}, 2008.

\bibitem{blanz2004exchanging}
Volker Blanz, Kristina Scherbaum, Thomas Vetter, and Hans-Peter Seidel.
\newblock Exchanging faces in images.
\newblock In {\em Computer Graphics Forum}, volume~23, pages 669--676. Wiley
  Online Library, 2004.

\bibitem{vggface2}
Qiong Cao, Li Shen, Weidi Xie, Omkar~M Parkhi, and Andrew Zisserman.
\newblock Vggface2: A dataset for recognising faces across pose and age.
\newblock {\em arXiv preprint arXiv:1710.08092}, 2017.

\bibitem{infogan}
Xi Chen, Xi Chen, Yan Duan, Rein Houthooft, John Schulman, Ilya Sutskever, and
  Pieter Abbeel.
\newblock {InfoGAN}: Interpretable representation learning by information
  maximizing generative adversarial nets.
\newblock In {\em NIPS}. 2016.

\bibitem{xception}
Francois Chollet.
\newblock Xception: Deep learning with depthwise separable convolutions.
\newblock In {\em Proceedings of the IEEE Conference on Computer Vision and
  Pattern Recognition}, pages 1251--1258, 2017.

\bibitem{arcface}
Jiankang Deng, Jia Guo, and Stefanos Zafeiriou.
\newblock Arcface: Additive angular margin loss for deep face recognition.
\newblock {\em arXiv preprint arXiv:1801.07698}, 2018.

\bibitem{dfgithub}
Faceswap.
\newblock Github project, https://github.com/deepfakes/faceswap.
\newblock 2017.

\bibitem{gan}
Ian Goodfellow, Jean Pouget-Abadie, Mehdi Mirza, Bing Xu, David Warde-Farley,
  Sherjil Ozair, Aaron Courville, and Yoshua Bengio.
\newblock Generative adversarial nets.
\newblock In {\em NIPS}. 2014.

\bibitem{4587369}
Ralph Gross, Latanya Sweeney, Fernando De~La Torre, and Simon Baker.
\newblock Semi-supervised learning of multi-factor models for face
  de-identification.
\newblock In {\em 2008 IEEE Conference on Computer Vision and Pattern
  Recognition}, 2008.

\bibitem{delving}
Kaiming He, Xiangyu Zhang, Shaoqing Ren, and Jian Sun.
\newblock Delving deep into rectifiers: Surpassing human-level performance on
  imagenet classification.
\newblock In {\em ICCV}, 2015.

\bibitem{resnet}
Kaiming He, Xiangyu Zhang, Shaoqing Ren, and Jian Sun.
\newblock Deep residual learning for image recognition.
\newblock In {\em CVPR}, 2016.

\bibitem{lfw}
Gary~B Huang, Manu Ramesh, Tamara Berg, and Erik Learned-Miller.
\newblock Labeled faces in the wild: A database for studying face recognition
  in unconstrained environments.
\newblock Technical report.

\bibitem{munit}
Xun Huang, Ming-Yu Liu, Serge Belongie, and Jan Kautz.
\newblock Multimodal unsupervised image-to-image translation.
\newblock In {\em ECCV}, 2018.

\bibitem{perceptual}
Justin Johnson, Alexandre Alahi, and Li Fei-Fei.
\newblock Perceptual losses for real-time style transfer and super-resolution.
\newblock In {\em ECCV}, 2016.

\bibitem{jourabloo2015attribute}
Amin Jourabloo, Xi Yin, and Xiaoming Liu.
\newblock Attribute preserved face deidentification.
\newblock In {\em In ICB}, 2015.

\bibitem{karras2017progressive}
Tero Karras, Timo Aila, Samuli Laine, and Jaakko Lehtinen.
\newblock Progressive growing of gans for improved quality, stability, and
  variation.
\newblock In {\em ICLR}, 2018.

\bibitem{kazemi2014one}
Vahid Kazemi and Josephine Sullivan.
\newblock One millisecond face alignment with an ensemble of regression trees.
\newblock In {\em Proceedings of the IEEE conference on computer vision and
  pattern recognition}, pages 1867--1874, 2014.

\bibitem{Kemelmacher-Shlizerman:2016:TP:2897824.2925871}
Ira Kemelmacher-Shlizerman.
\newblock Transfiguring portraits.
\newblock {\em ACM Trans. Graph.}, 35(4), 2016.

\bibitem{discogan}
Taeksoo Kim, Moonsu Cha, Hyunsoo Kim, Jungkwon Lee, and Jiwon Kim.
\newblock Learning to discover cross-domain relations with generative
  adversarial networks.
\newblock {\em arXiv preprint arXiv:1703.05192}, 2017.

\bibitem{dlib}
Davis~E King.
\newblock Dlib-ml: A machine learning toolkit.
\newblock {\em Journal of Machine Learning Research}, 10(Jul):1755--1758, 2009.

\bibitem{adam}
Kingma, Diederik P., and Jimmy Ba.
\newblock Adam: A method for stochastic optimization.
\newblock In {\em ICLR}, 2016.

\bibitem{korshunova2017fast}
Iryna Korshunova, Wenzhe Shi, Joni Dambre, and Lucas Theis.
\newblock Fast face-swap using convolutional neural networks.
\newblock In {\em The IEEE International Conference on Computer Vision}, 2017.

\bibitem{pubfig}
Neeraj Kumar, Alexander~C Berg, Peter~N Belhumeur, and Shree~K Nayar.
\newblock Attribute and simile classifiers for face verification.
\newblock In {\em CVPR}, 2009.

\bibitem{fader}
Guillaume Lample et~al.
\newblock Fader networks: Manipulating images by sliding attributes.
\newblock In {\em NIPS}, 2017.

\bibitem{drit}
Hsin-Ying Lee, Hung-Yu Tseng, Jia-Bin Huang, Maneesh Singh, and Ming-Hsuan
  Yang.
\newblock Diverse image-to-image translation via disentangled representations.
\newblock In {\em The European Conference on Computer Vision (ECCV)}, September
  2018.

\bibitem{unit}
Ming-Yu Liu, Thomas Breuel, and Jan Kautz.
\newblock Unsupervised image-to-image translation networks.
\newblock In {\em NIPS}. 2017.

\bibitem{celeba}
Ziwei Liu, Ping Luo, Xiaogang Wang, and Xiaoou Tang.
\newblock Deep learning face attributes in the wild.
\newblock In {\em ICCV}, 2015.

\bibitem{makhzani2015adversarial}
Alireza Makhzani, Jonathon Shlens, Navdeep Jaitly, Ian Goodfellow, and Brendan
  Frey.
\newblock Adversarial autoencoders.
\newblock {\em arXiv preprint arXiv:1511.05644}, 2015.

\bibitem{lsgan}
Xudong Mao, Qing Li, Haoran Xie, Raymond~YK Lau, Zhen Wang, and Stephen~Paul
  Smolley.
\newblock Least squares generative adversarial networks.
\newblock In {\em ICCV}, 2017.

\bibitem{meden2017face}
Bla{\v{z}} Meden, Refik~Can Mall{\i}, Sebastjan Fabijan, Haz{\i}m~Kemal Ekenel,
  Vitomir {\v{S}}truc, and Peter Peer.
\newblock Face deidentification with generative deep neural networks.
\newblock {\em IET Signal Processing}, 11(9):1046--1054, 2017.

\bibitem{1377174}
Elaine~M Newton, Latanya Sweeney, and Bradley Malin.
\newblock Preserving privacy by de-identifying face images.
\newblock {\em IEEE transactions on Knowledge and Data Engineering},
  17(2):232--243, 2005.

\bibitem{newton2005preserving}
Elaine~M Newton, Latanya Sweeney, and Bradley Malin.
\newblock Preserving privacy by de-identifying face images.
\newblock {\em IEEE transactions on Knowledge and Data Engineering},
  17(2):232--243, 2005.

\bibitem{nirkin2017face}
Yuval Nirkin, Iacopo Masi, Anh~Tuan Tran, Tal Hassner, and Gerard Medioni.
\newblock On face segmentation, face swapping, and face perception.
\newblock {\em arXiv preprint arXiv:1704.06729}, 2017.

\bibitem{oh2017adversarial}
Seong~Joon Oh, Mario Fritz, and Bernt Schiele.
\newblock Adversarial image perturbation for privacy protection a game theory
  perspective.
\newblock In {\em 2017 IEEE International Conference on Computer Vision
  (ICCV)}, pages 1491--1500. IEEE, 2017.

\bibitem{gbu}
P~Jonathon Phillips, J~Ross Beveridge, Bruce~A Draper, Geof Givens, Alice~J
  O'Toole, David~S Bolme, Joseph Dunlop, Yui~Man Lui, Hassan Sahibzada, and
  Samuel Weimer.
\newblock An introduction to the good, the bad, \& the ugly face recognition
  challenge problem.
\newblock In {\em Automatic Face \& Gesture Recognition}, 2011.

\bibitem{dcgan}
Alec Radford, Luke Metz, and Soumith Chintala.
\newblock Unsupervised representation learning with deep convolutional
  generative adversarial networks.
\newblock {\em arXiv preprint arXiv:1511.06434}, 2015.

\bibitem{unet}
Olaf Ronneberger, Philipp Fischer, and Thomas Brox.
\newblock U-net: Convolutional networks for biomedical image segmentation.
\newblock In {\em International Conference on Medical image computing and
  computer-assisted intervention}, pages 234--241. Springer, 2015.

\bibitem{roessler2018faceforensics}
Andreas R\"ossler, Davide Cozzolino, Luisa Verdoliva, Christian Riess, Justus
  Thies, and Matthias Nie{\ss}ner.
\newblock Face{F}orensics: A large-scale video dataset for forgery detection in
  human faces.
\newblock {\em arXiv}, 2018.

\bibitem{gantricks}
Tim Salimans, Ian~J. Goodfellow, Wojciech Zaremba, Vicki Cheung, Alec Radford,
  and Xi Chen.
\newblock Improved techniques for training gans.
\newblock {\em arXiv preprint arXiv:1606.03498}, 2016.

\bibitem{6859758}
Branko Samarzija and Slobodan Ribaric.
\newblock An approach to the de-identification of faces in different poses.
\newblock In {\em 2014 37th International Convention on Information and
  Communication Technology, Electronics and Microelectronics (MIPRO)}, pages
  1246--1251. IEEE, 2014.

\bibitem{schroff2015facenet}
Florian Schroff, Dmitry Kalenichenko, and James Philbin.
\newblock Facenet: A unified embedding for face recognition and clustering.
\newblock In {\em Proceedings of the IEEE conference on computer vision and
  pattern recognition}, pages 815--823, 2015.

\bibitem{sun2018natural}
Qianru Sun, Liqian Ma, Seong Joon~Oh, Luc Van~Gool, Bernt Schiele, and Mario
  Fritz.
\newblock Natural and effective obfuscation by head inpainting.
\newblock In {\em Proceedings of the IEEE Conference on Computer Vision and
  Pattern Recognition}, pages 5050--5059, 2018.

\bibitem{sun2018hybrid}
Qianru Sun, Ayush Tewari, Weipeng Xu, Mario Fritz, Christian Theobalt, and
  Bernt Schiele.
\newblock A hybrid model for identity obfuscation by face replacement.
\newblock In {\em Proceedings of the European Conference on Computer Vision
  (ECCV)}, pages 553--569, 2018.

\bibitem{02200}
Yaniv Taigman, Adam Polyak, and Lior Wolf.
\newblock Unsupervised cross-domain image generation.
\newblock In {\em International Conference on Learning Representations (ICLR)},
  2017.

\bibitem{thies2016face2face}
Justus Thies, Michael Zollhofer, Marc Stamminger, Christian Theobalt, and
  Matthias Nie{\ss}ner.
\newblock Face2face: Real-time face capture and reenactment of rgb videos.
\newblock In {\em Proceedings of the IEEE Conference on Computer Vision and
  Pattern Recognition}, pages 2387--2395, 2016.

\bibitem{ulyanov2016texture}
Dmitry Ulyanov, Vadim Lebedev, Victor Lempitsky, et~al.
\newblock Texture networks: Feed-forward synthesis of textures and stylized
  images.
\newblock In {\em ICML}, 2016.

\bibitem{ulyanov2016instance}
Dmitry Ulyanov, Andrea Vedaldi, and Victor Lempitsky.
\newblock Instance normalization: The missing ingredient for fast stylization.
\newblock {\em arXiv preprint arXiv:1607.08022}, 2016.

\bibitem{ppgan}
Yifan Wu, Fan Yang, and Haibin Ling.
\newblock Privacy-protective-gan for face de-identification.
\newblock {\em arXiv preprint arXiv:1806.08906}, 2018.

\bibitem{xie2018feature}
Cihang Xie et~al.
\newblock Feature denoising for improving adversarial robustness.
\newblock {\em arXiv preprint arXiv:1812.03411}, 2018.

\bibitem{dualgan}
Zili Yi, Hao Zhang, Ping Tan, and Minglun Gong.
\newblock {DualGAN}: Unsupervised dual learning for image-to-image translation.
\newblock {\em arXiv preprint arXiv:1704.02510}, 2017.

\bibitem{mixup}
Hongyi Zhang, Moustapha Cisse, Yann~N Dauphin, and David Lopez-Paz.
\newblock mixup: Beyond empirical risk minimization.
\newblock {\em arXiv preprint arXiv:1710.09412}, 2017.

\end{thebibliography}
}

\clearpage
\FloatBarrier
\appendix

\begin{figure}
\centering
  \includegraphics[width=0.96\linewidth]{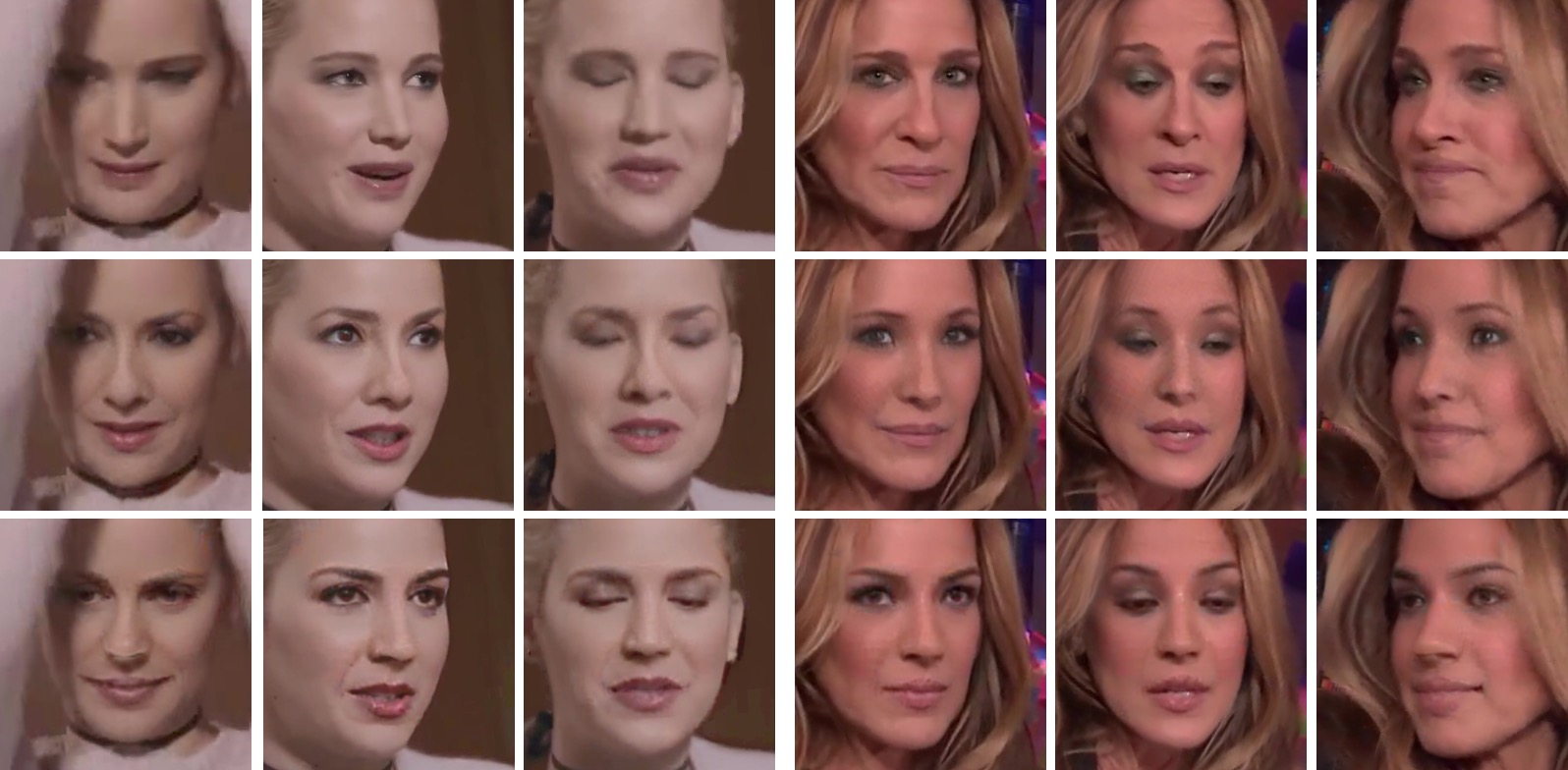}
   \caption{No face-descriptor ablation study. Source (row 1), our model (row 2), and no face-descriptor (row 3), resulting in lower quality results, with noticeable artifacts in the rendered identity.}
\label{fig:z_abl}
\centering
 \begin{tabular}{@{}c@{~}c@{~}c@{~}c} 
  \includegraphics[height=5.5cm]{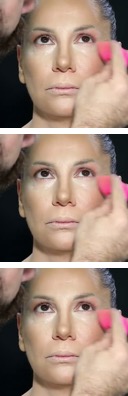} & 
  \includegraphics[height=5.5cm]{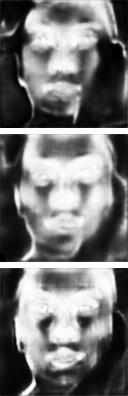} &
  \includegraphics[height=5.5cm]{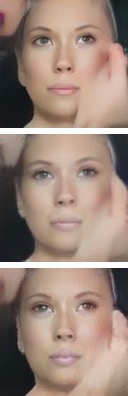} & 
  \includegraphics[height=5.5cm]{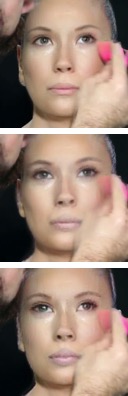} \\
  (a) & (b) & (c) & (d)
  \end{tabular}
  
\caption{Mask regularization ablation study and mask outputs. (a) Source, (b) mask, (c) raw output, (d) masked output. Compared to our model (row 1), the effects of not minimizing the mask norm ($\alpha_4=0$, row 2) can be observed as occlusions (hand, pink element, upper-left face) not handled well, and excessive face regions taken from the rendered image, resulting in distortions. No mask derivative regularization ($\alpha_5=0$, row 3) effects can be seen as high-frequency patterns generated by the mask and output frame.}

\label{fig:ablation1}

\end{figure}

\section{Ablation Analyses}
{\color{black}
\subsection{General Ablation Analysis}
}

\begin{figure*}[t]
  \centering
  \resizebox{1.0\textwidth}{!}{
  \begin{tabular}{@{}c@{~}c@{~}c@{~}c@{~}c@{~}c@{~}c@{}}
  \includegraphics[width=1.89cm, height=2.24cm]{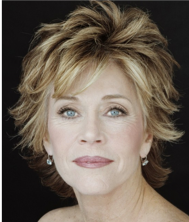} & \includegraphics[width=1.89cm, height=2.24cm]{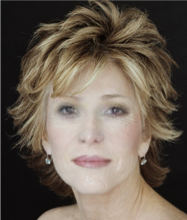} & \includegraphics[width=1.89cm, height=2.24cm]{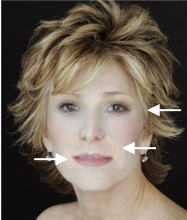} & \includegraphics[width=1.89cm, height=2.24cm]{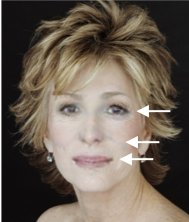} & \includegraphics[width=1.89cm, height=2.24cm]{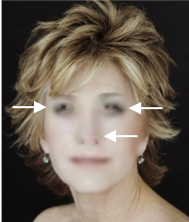} & 
  \includegraphics[width=1.89cm, height=2.24cm]{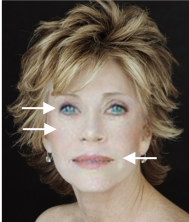} & \includegraphics[width=1.89cm, height=2.24cm]{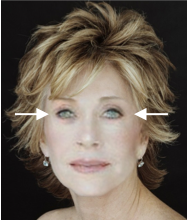} 
   \\
    \includegraphics[width=1.89cm, height=2.24cm]{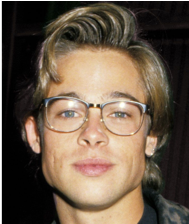} & \includegraphics[width=1.89cm, height=2.24cm]{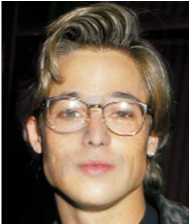} & \includegraphics[width=1.89cm, height=2.24cm]{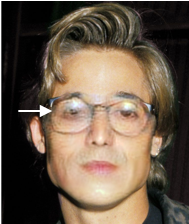} & 
    \includegraphics[width=1.89cm, height=2.24cm]{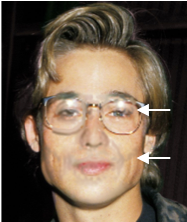} & \includegraphics[width=1.89cm, height=2.24cm]{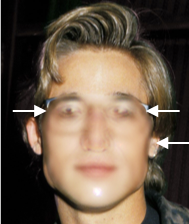} & \includegraphics[width=1.89cm, height=2.24cm]{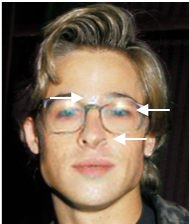} & 
    \includegraphics[width=1.89cm, height=2.24cm]{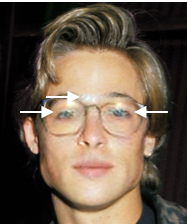}\\
  (a) & (b) & (c) & (d) & (e) & (f) & (g) 
  \end{tabular} 
  }
  \caption{An ablation study. (a) Source image. (b) Our result. (c)--(g) variants, see text for details.}
  \label{fig:ablation}
\end{figure*}

An ablation analysis is shown in Fig.~\ref{fig:ablation}. {\color{black} In order to efficiently run multiple-models, it is done with the low-res architecture.} The various options include: a no-mask option, a partial adversarial loss that applies only to the masked output and not to the raw output, training without the gradual increase of $\lambda$, and an attempt to incorporate an additional output with a lower resolution to be taken into account, as part of the compound loss. All of these ablation experiments were conducted on the lower-resolution model. 

The following description of methods and the associated artifacts correspond to the columns of Fig.~\ref{fig:ablation}:  (c) No mask. {\em Bad face edge, glasses occlusion handled poorly.} (d) Adversarial loss on masked output only. {\em Various artifacts, e.g., around the right eye, one can also observe green stripes near the mouth.} (e) No gradual increment of $\lambda$. {\em Collapse into unnatural blurred face}. (f) Lower resolution output added for the compound loss. {\em Weak de-id, checkerboard pattern near the center of the face, and when handling occlusions.} (g) Weak $\lambda$, adversarial loss on masked output only. {\em Weak de-identification, artifacts near the eyes and eyebrows}.

\begin{figure*}[t]  
  \includegraphics[width=.9\linewidth]{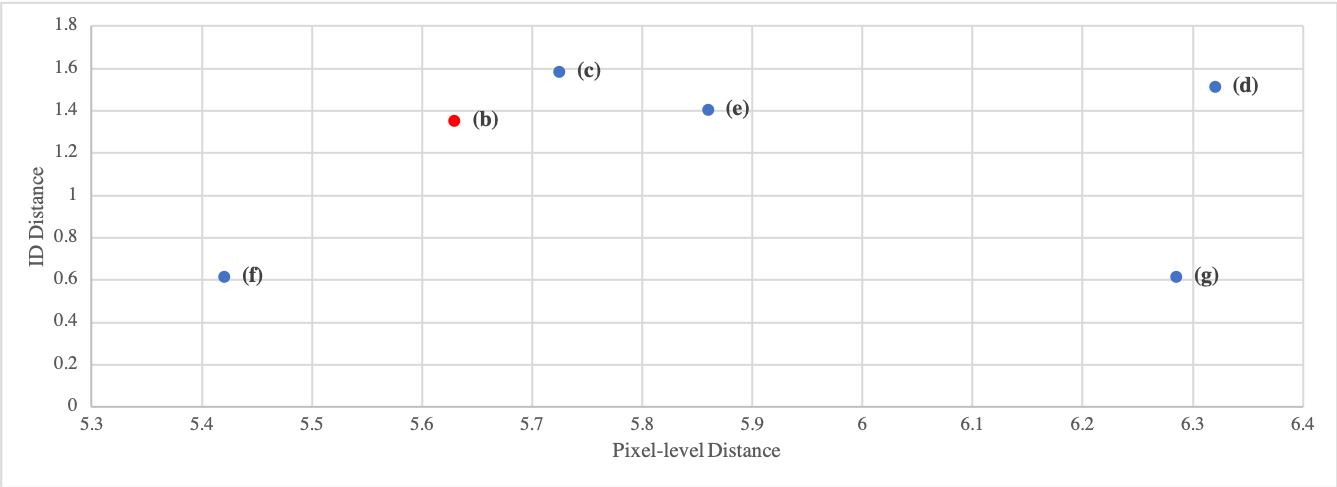}
  \caption{The mean pixel-level distance vs. the mean ID distance. The first should be low, while the second should be high. Shown are the methods of each column in Fig.\ref{fig:ablation}(b)--(g).}
  \label{fig:ablationplot}
\end{figure*}

A numerical analysis plot of the ablation study with the same options is provided in Fig.~\ref{fig:ablationplot}. Each method is evaluated along two axes of comparison between the input image and the output image: on the x-axis we show the difference in appearance as measured by the L1 norm between the images; the y-axis shows the difference in ID, as computed by the L1 norm between the VGGFace2 representation of the two images. The plot shows mean results obtained for our method (marked (b) to match the columns of  Fig.~\ref{fig:ablation}) and the various ablation methods (marked (c)--(g)). As can be seen, our method maintains image similarity and also has a difference in ID that is similar or larger than any other method, with the exception of the method marked as (c). This is expected, since this variant is the mask-less one, which does not blend-in the original image. Variant (f) is considerably more similar to the original image on both axes, since the de-ID performed is very weak with this variant.

{\color{black}
\subsection{Face-Descriptor Ablation Analysis}
A face-descriptor-specific ablation analysis is provided to emphasize its necessity in Fig.~\ref{fig:z_abl}. The face-descriptor is highly motivating for the decoder to use, otherwise, minimizing the high-level perceptual loss ($l_{1\times1}$) would be more challenging, as can be seen in Fig.~\ref{fig:z_abl}. For each source image (row 1), our model result (row 2) can be seen to produce higher-quality results with less artifacts, compared to the model that lacks a face-descriptor concatenated to the latent space (row 3).
In the results of the third row, the face descriptor is not concatenated to the z embedding, but still used in the perceptual loss. 
}

{\color{black}
\subsection{Mask Regularization Ablation Analysis}
The mask regularization parameters $\alpha_{4,5}$ importance can be observed in Fig.~\ref{fig:ablation1}. They assist in dealing with occlusions, and handling irrelevant regions, that can be taken from the source image, rather than generated (e.g. regions that are not related to the generated face, teeth, etc.). $\alpha_{4}$ keeps the mask minimal, i.e. blending maximal regions from the source image, rather than the generated one. By avoiding excessive blending of generated regions, less artifacts are apparent on the final output (as observed in row 2). $\alpha_{4}$ keeps the mask smooth, by penalizing mask derivatives. This can be seen to reduce high-frequency patterns, (as observed in row 3). 
}

\section{Additional Comparison with Previous Methods}
\begin{figure*}[t]
  \centering
  \begin{tabular}{ccc@{~}c@{~}c@{~}c@{~}c@{~}c@{}}
  \includegraphics[width=1.8cm, height=6.5165cm]{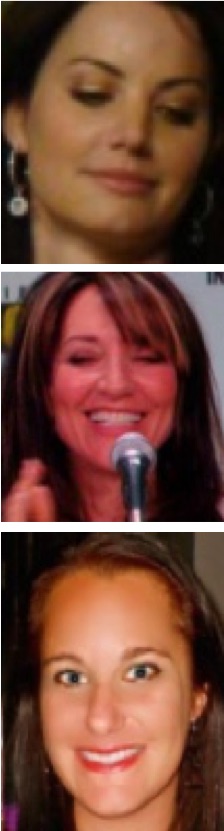} & \includegraphics[width=1.8cm, height=6.5165cm]{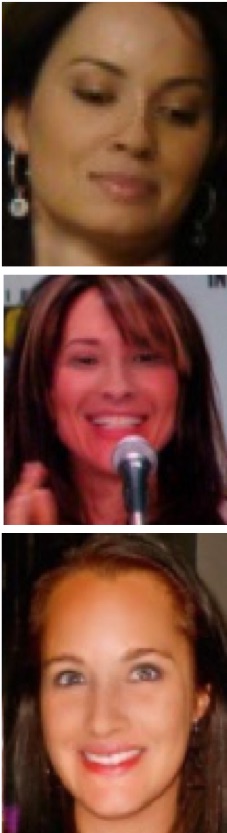} & \includegraphics[width=1.8cm, height=6.5165cm]{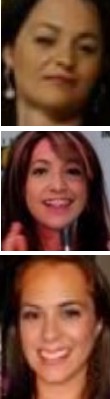} & \includegraphics[width=1.8cm, height=6.5165cm]{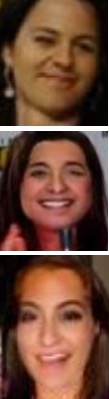} & \includegraphics[width=1.8cm, height=6.5165cm]{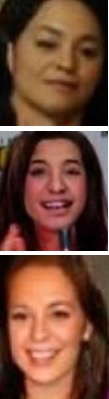} & \includegraphics[width=1.8cm, height=6.5165cm]{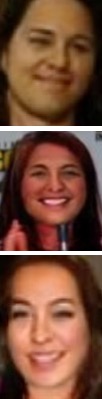} & \includegraphics[width=1.8cm, height=6.5165cm]{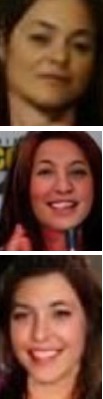} & \includegraphics[width=1.8cm, height=6.5165cm]{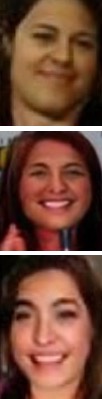}
   \\
  (a) & (b) & (c) & (d) & (e) & (f) & (g) & (h)
  \end{tabular} 
  \caption{A full set of results for the comparison with the work of~\cite{sun2018hybrid}. (a) Source image, (b) our generated output, (c-h) generated outputs for~\cite{sun2018hybrid} of different models. The work of~\cite{sun2018hybrid} provides several models that provide for different levels of de-identification. As can be seen, models that gain a rate of $<50\%$ of head obfuscation effectiveness by machine recognizers, provide less natural faces.}
  \label{fig:hybrid_supp}
\end{figure*}

We provide an extensive comparison with the work of~\cite{sun2018hybrid}. In the paper we only included one of~\cite{sun2018hybrid} generated outputs: the sample shown was the first output that gains <50\% of recognition rates by an automatic face recognition algorithm, according to~\cite{sun2018hybrid}. The work of~\cite{sun2018hybrid} provides several models for different levels of de-identification. In Fig.~\ref{fig:hybrid_supp}, we present all faces from~\cite{sun2018hybrid}. The reported recognition rate itself is given in Tab.~\ref{tab:hybrid_comp_supp}. 

As can be seen in the results of~\cite{sun2018hybrid}, the less recognizable the identity is, the less natural the face is. Note that: (1) our model provides for much stronger de-identification results, with the rank typically in the thousands, out of a dataset of 54,000 persons, as reported in the experiments section. (2) all models of the baseline method produce low resolution outputs ($64\times64$) compared to our model's much higher resolution ($256\times256$). 


\clearpage

\begin{table*}[t]
\begin{center}
\vspace{-7.5cm}
\begin{tabular}[b]{lcccccc}
    \toprule
    Row | Column & (c) & (d) & (e) & (f) & (g) & (h)\\
    \midrule
Row 1&	70.8\% & 47.6\% & 36.6\% & 18.0\% & 22.5\% & 7.1\% \\
Row 2&	59.9\% & 26.3\% & 25.8\% & 12.7\% & 15.7\% & 7.2\% \\
Row 3&	59.9\% & 26.3\% & 25.8\% & 12.7\% & 15.7\% & 7.2\% \\
    \bottomrule
  \end{tabular} 
\end{center}
  \caption{Head obfuscation effectiveness for~\cite{sun2018hybrid}: recognition rates of machine recognizers (lower is better), as provided by~\cite{sun2018hybrid}}
  \label{tab:hybrid_comp_supp}
\end{table*}

\end{document}